\documentclass[12pt,twocolumn]{article}

\usepackage[utf8]{inputenc}
\usepackage[T1]{fontenc}
\usepackage{graphicx}
\usepackage{sectsty}
\usepackage{titlesec}
\usepackage{enumitem}
\usepackage{fancyhdr}
\usepackage{geometry}
\usepackage{booktabs}
\usepackage{longtable}
\usepackage{array}
\usepackage{hyperref}
\usepackage{parskip}
\usepackage[numbers]{natbib}
\usepackage{tabularx}
\usepackage{microtype}
\usepackage{enumitem}
\usepackage{amsmath}
\usepackage{amssymb}
\usepackage{siunitx}
\usepackage{float}
\usepackage{tikz}
\usepackage{svg}

\usetikzlibrary{positioning}

\usepackage{listings}
\titleformat{\section}{\Large\bfseries}{\thesection}{1em}{}
\titleformat{\subsection}{\large\bfseries}{\thesubsection}{1em}{}

\title{
    \Huge \textbf{Final Project Report}\\[1em]
    \Large \textbf{Whisper: Courtside Edition}
}
\author{%
  \textbf{Yonathan Ron}\\
  Reichman University\\
  \and
  \textbf{Shiri Gilboa}\\
  Reichman University\\
}
\date{\today}

\begin{document}

% Create a full-width title block before the two-column content starts.
\twocolumn[{
\centering

{\Large Whisper: Courtside Edition}\\[1em]
{\large Enhancing ASR Performance Through LLM-Driven Context Generation}\\[1.2em]

{\large
Yonathan Ron \quad Shiri Gilboa \quad Tammuz Dubnov
}\\[0.6em]

{\normalsize
Reichman University
}\\[1em]

{\small \today}
\begin{flushleft}
\begin{abstract}
\textbf{Abstract:} Domain-specific speech remains a persistent challenge for automatic speech recognition (ASR), even for state-of-the-art systems like OpenAI's Whisper. We introduce \textit{Whisper: Courtside Edition}, a novel multi-agent large language model (LLM) pipeline that enhances Whisper transcriptions without retraining. The pipeline intercepts Whisper’s initial transcript, applies specialized LLM agents for domain context identification, named entity recognition, and jargon detection, and generates compact prompts that guide Whisper’s decoder. Evaluated on 421 NBA basketball commentary segments—a domain characterized by dense proper nouns and technical terminology—our best pipeline achieves a statistically significant 17.0\% relative reduction in word error rate (WER; from 0.217 to 0.180, $p < 0.001$). Improvements are observed in 40.1\% of segments with degradation in only 7.1\%, substantially outperforming direct transcript post-editing. These results demonstrate that prompt-based augmentation can deliver scalable domain adaptation for ASR, offering a practical alternative to costly model fine-tuning.
\end{abstract}
\end{flushleft}
\vspace{1em}
}]
\clearpage
\twocolumn

\section{Introduction}

Modern automatic speech recognition systems like OpenAI's Whisper\cite{radford2023} have revolutionized speech-to-text conversion through Transformer architectures trained on massive multilingual datasets. While these models achieve near human-level accuracy on general-domain speech, they exhibit systematic failures on specialized content containing domain-specific terminology, proper nouns, and rapid contextual speech patterns that fall outside their training distribution.

\textbf{Problem Motivation:} Domain-specific ASR challenges are particularly acute in scenarios such as sports commentary, medical dictation, and technical presentations. For instance, in NBA basketball commentary—our focus domain—transcription errors frequently involve player name mangling ("Giannis Antetokounmpo" → "yanis anteto kumbo"), basketball terminology corruption ("pick and roll" → "picker roll"), and context-dependent misinterpretations. These errors significantly reduce transcript utility, as semantic meaning can be completely altered by incorrect proper nouns or technical terms.

\textbf{Our Contribution:} We present a multi-agent LLM pipeline that intercepts Whisper's initial transcript and generates domain-aware prompts to guide a second transcription pass. Rather than post-processing corrections, our approach leverages Whisper's built-in prompting mechanism to inject contextual knowledge during decoding, allowing the ASR model to reconcile acoustic evidence with domain expertise. The key contributions include: (1) A novel multi-agent architecture combining topic classification, named entity correction, and jargon identification; (2) Comprehensive statistical evaluation demonstrating significant improvements over baseline and alternative approaches; (3) Detailed error taxonomy and failure mode analysis; and (4) A generalizable framework requiring only domain knowledge bases and LLM access, avoiding costly model retraining.
\section{Related Work}

\textbf{Domain Adaptation in ASR:}  
Traditional domain adaptation strategies for ASR include acoustic model fine-tuning and language model adaptation, which adapt models to domain-specific data but require expensive retraining and large annotated corpora \cite{yu2013, bellegarda2004}. More recently, contextual biasing methods have been proposed for end-to-end ASR, where external vocabulary lists or keyword spotters are injected during decoding to steer recognition toward domain-relevant terms \cite{zhao2019}. For example, CB-Whisper \cite{li2024cb} applies open-vocabulary keyword spotting to bias Whisper outputs, but its approach still involves model modification and retraining, limiting scalability. By contrast, our pipeline achieves domain-specific improvements without retraining, relying only on intelligent prompt generation.

\textbf{LLM-Enhanced ASR:}  
Large language models (LLMs) have recently been explored for speech transcription enhancement. Early efforts focus on post-processing, where an LLM edits ASR transcripts to fix spelling, casing, or grammar \cite{suh2024}. However, these approaches cannot recover information that Whisper misheard, since the LLM has no access to the audio. Other works, such as Zellers et al. \cite{zellers2024}, demonstrate the use of LLM-generated prompts to improve named entity recognition or context priming, but they are evaluated in narrow settings and lack systematic, domain-scale assessment. Our work advances this line by showing that coordinated prompt generation across multiple error types (names, jargon, context) can deliver robust, domain-wide improvements.

\textbf{Multi-Agent Systems:}  
Recent advances in multi-agent LLM systems highlight the benefits of specialization and collaboration. Frameworks such as AutoGPT and HuggingFace Transformers Agents demonstrate that decomposing tasks into subtasks handled by different agents can outperform monolithic approaches \cite{guo2024}. Inspired by this paradigm, our system assigns Whisper-augmentation subtasks to distinct agents (topic detection, named entity normalization, jargon handling, decision filtering), enabling precise targeting of error modalities in ASR. To our knowledge, this is the first application of a multi-agent LLM architecture to enhance speech recognition via Whisper’s prompt mechanism.

\section{Problem Analysis and Dataset}

\subsection{NBA Commentary Dataset}

Our primary evaluation corpus comprises 421 segments from NBA basketball commentary, selected to represent diverse challenges: multiple broadcasters, teams, game contexts, accents, and acoustic conditions with frequent crowd noise and speaker overlap. Segments range from 10--30 seconds (median 15s) to balance contextual sufficiency with processing efficiency.

\textbf{Collection Methodology:}
We gather video metadata and select clips using stratified sampling over channel/source, team matchup, game situation (e.g., clutch moments), and season-year to avoid topical skew. Segments are cut at natural commentary boundaries and filtered to exclude music-only interludes or on-court interviews. To reduce topical leakage across splits, we enforce disjointness by game and broadcast team.

\textbf{Annotation and Expert Tagging:}
All segments were transcribed and normalized by \emph{domain-expert} annotators with basketball commentary experience. Experts resolve rapid elisions, handle overlapped speech, and canonicalize proper nouns (players/coaches/referees) to official spellings. The guideline includes: (i) orthography normalization (numbers, casing, punctuation), (ii) bracketed tags for clear overlaps and non-lexical events, and (iii) entity canonicalization against official rosters.

\textbf{Data Splits:} To ensure evaluation integrity, we enforced disjoint splits by game and broadcast team, preventing information leakage between training context and test evaluation.

\textbf{Run Transcription:}
Before conducting a detailed error analysis, we established a baseline by running OpenAI's \texttt{whisper-medium.en} model on the test set. The resulting average Word Error Rate (WER) was \textbf{0.217}, reflecting the inherent difficulty of transcribing rapid, jargon-heavy basketball commentary under varied acoustic conditions.

\subsection{Error Taxonomy}

Analysis of Whisper's baseline performance revealed systematic error patterns:\\

\textbf{Phonetic Substitution Errors - Names (35\% of errors):} Names replaced with phonetically similar common words ("Kristaps Porzingis" → "Christmas Por Zingas").\\

\textbf{Domain Jargon Errors (28\% of errors):} Basketball terminology corrupted ("pick and roll" → "picker roll", "alley-oop" → "alley oops").\\

\textbf{Accent/Dialect Sensitivity (22\% of errors):} Non-standard pronunciations causing systematic misrecognition, particularly for international player names.\\

\textbf{Rapid Speech Segmentation (15\% of errors):} Word boundary errors and omissions during excited commentary segments.

This taxonomy directly informed our multi-agent pipeline design, with specialized agents targeting each error category.

\section{Methodology}

\subsection{Multi-Agent Pipeline Architecture}

Our approach employs six specialized LLM agents operating on Whisper's initial transcript: described in detail in Appendix~\ref{appendix:prompts}.

\textbf{Topic Classification Agent:} Identifies domain context (e.g., "NBA basketball commentary") to activate appropriate knowledge bases and set processing expectations.

\textbf{Named Entity Recognition Agent:} Extracts person names from the transcript and applies fuzzy matching (Levenshtein distance) against comprehensive NBA player rosters. Similarity thresholds were empirically optimized to $\tau$ = 0.75 for precision--recall balance.

\textbf{Jargon Extraction Agent:} Identifies domain-specific terminology using keyword extraction (RAKE, YAKE) combined with basketball glossary lookup. Statistical significance testing ensures extracted terms exceed random occurrence.

\textbf{Decision Filtering Logic:} Critical validation agents prevent over-correction by verifying proposed names represent actual corrections, ensuring jargon additions improve rather than degrade semantic coherence, and maintaining prompt length within Whisper's 224-token limit.

\textbf{Candidate selector} Resolves conflicts across sources by attaching per-item scores and pruning to a compact shortlist.

\textbf{Sentence builder} Converts the selected topic, names, and jargon into one concise sentence that reads naturally and places the \emph{highest-value tokens near the end}, exploiting Whisper’s behavior of considering only the final $\le 224$ tokens.

\subsection{Whisper Initial Prompt Mechanism}

\subsubsection*{Background on Whisper Initial Prompt}
\begin{figure*}[t]
    \centering
    \includegraphics[width=\textwidth]{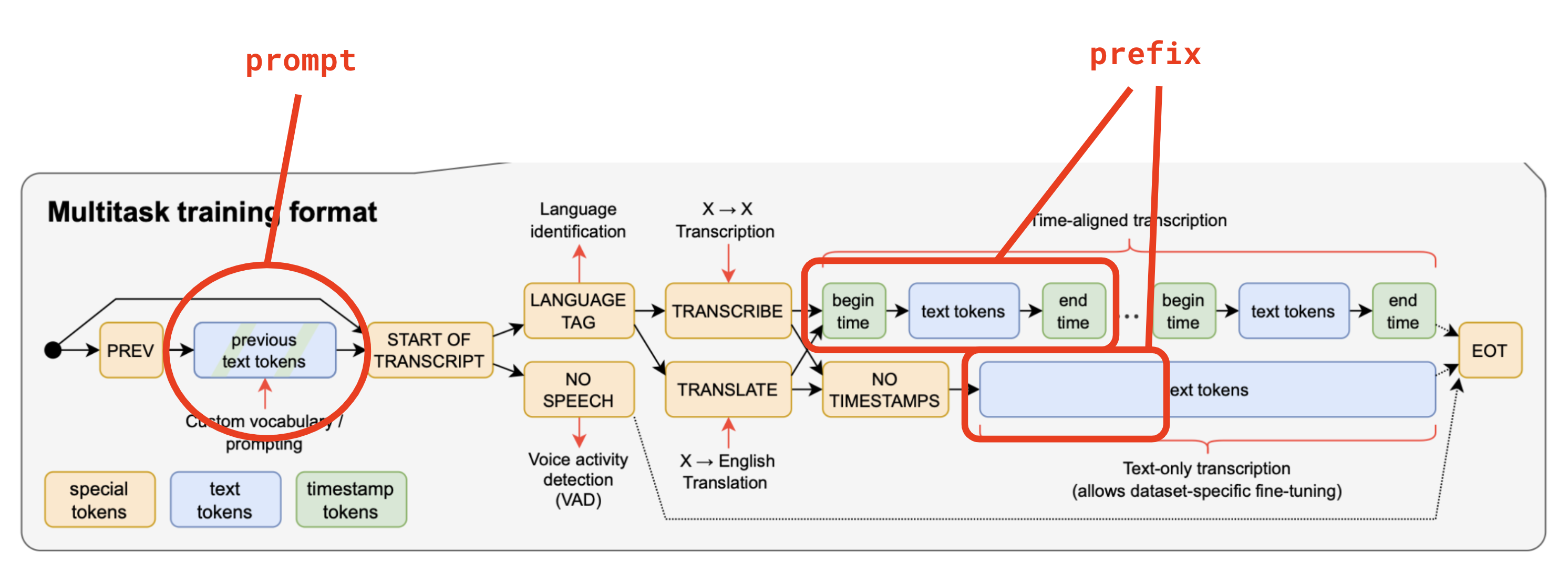}
    \caption{Original Whisper decoding format (adapted from the official documentation), 
    annotated to highlight where the \texttt{initial\_prompt} (red circle, previous text tokens) 
    and prefix tokens (red boxes, time-aligned text tokens) enter the model’s input stream. 
    In our work, we leverage the initial prompt channel to inject domain-specific context, 
    enabling Whisper to bias its decoding toward correct player names and basketball jargon.}
    \label{fig:whisper_initial_prompt}
\end{figure*}

Whisper supports an \texttt{initial\_prompt} parameter that allows contextual text to be prepended to the decoder’s input sequence at inference time. According to the official Whisper prompting guide\cite{openai_whisper_prompting}, this feature can bias transcription toward “preferred vocabulary, spellings, or styles” by providing relevant text before decoding begins. Only the last $\leq 224$ tokens of the prompt are consumed, which means later tokens exert greater influence than earlier ones. In practice, effective prompts must therefore be compact, prioritize rare or domain-specific words, and place the highest-value tokens near the end. Unlike GPT-style prompting, which operates over free-form reasoning, Whisper prompting is strictly about lexical bias during decoding.
Figure~\ref{fig:whisper_initial_prompt} illustrates the underlying mechanism: 
Whisper prepends prompt tokens into the decoder’s input sequence (red circle), 
while also supporting prefix tokens (red boxes) for continuation tasks. 
Our pipeline exploits the prompt pathway to bias decoding toward domain-specific vocabulary.

\subsubsection*{Relevance to Our Pipeline}
This mechanism is the key enabler of our approach. Instead of applying corrections after Whisper has already produced its transcript, we use the \texttt{initial\_prompt} channel to steer the model during transcription itself. Our multi-agent pipeline analyzes Whisper’s first-pass transcript and constructs a short natural-sentence prompt that encodes three elements: (i) topic/domain context, (ii) canonical player names, and (iii) basketball jargon. Injecting this sentence into Whisper’s second decoding pass biases the model toward correct recognition of proper nouns and domain-specific terminology while still reconciling the text with the acoustic evidence. Without the \texttt{initial\_prompt} feature, our system would be like post hoc editing; with it, we achieve a true context-aware ASR enhancement that directly leverages Whisper's decoder.

\subsection{Pipeline Variants}

We implemented four progressively sophisticated pipelines for controlled evaluation, all pipelines utilized few-shot prompting. \\

\textbf{P1 (Topic-Only):} The simplest pipeline, which extracts a short topic label from the first-pass transcript and uses it directly as the \texttt{initial\_prompt}. This provides broad domain context (e.g., “NBA basketball commentary”) but no specific vocabulary. While sometimes effective for steering style, it often fails to resolve proper noun or jargon errors. \\

\textbf{P2 (LLM Post-Correction):} Instead of using Whisper’s prompt mechanism, this pipeline relies on GPT-4o to directly edit the first-pass transcript. The LLM is instructed to act as a careful proofreader, fixing spelling and formatting errors. However, because the LLM has no access to the audio signal, it cannot reliably correct misheard names or jargon, and may even introduce ungrounded substitutions. This serves as a useful post-processing baseline, but one with clear limitations. \\

\textbf{P3 (Names-Enhanced):} Building on P1, this pipeline augments the topic prompt with candidate player names detected by a Named Entity Recognition (NER) agent. The names are normalized to official spellings and injected into the prompt, biasing Whisper toward producing correct proper nouns. This significantly improves name-recognition accuracy, but without additional safeguards, it can lead to overcorrections (e.g., inserting a name not actually spoken). \\

\begin{figure*}[p]
    \centering
    \includegraphics[width=\textwidth,height=\textheight,keepaspectratio]{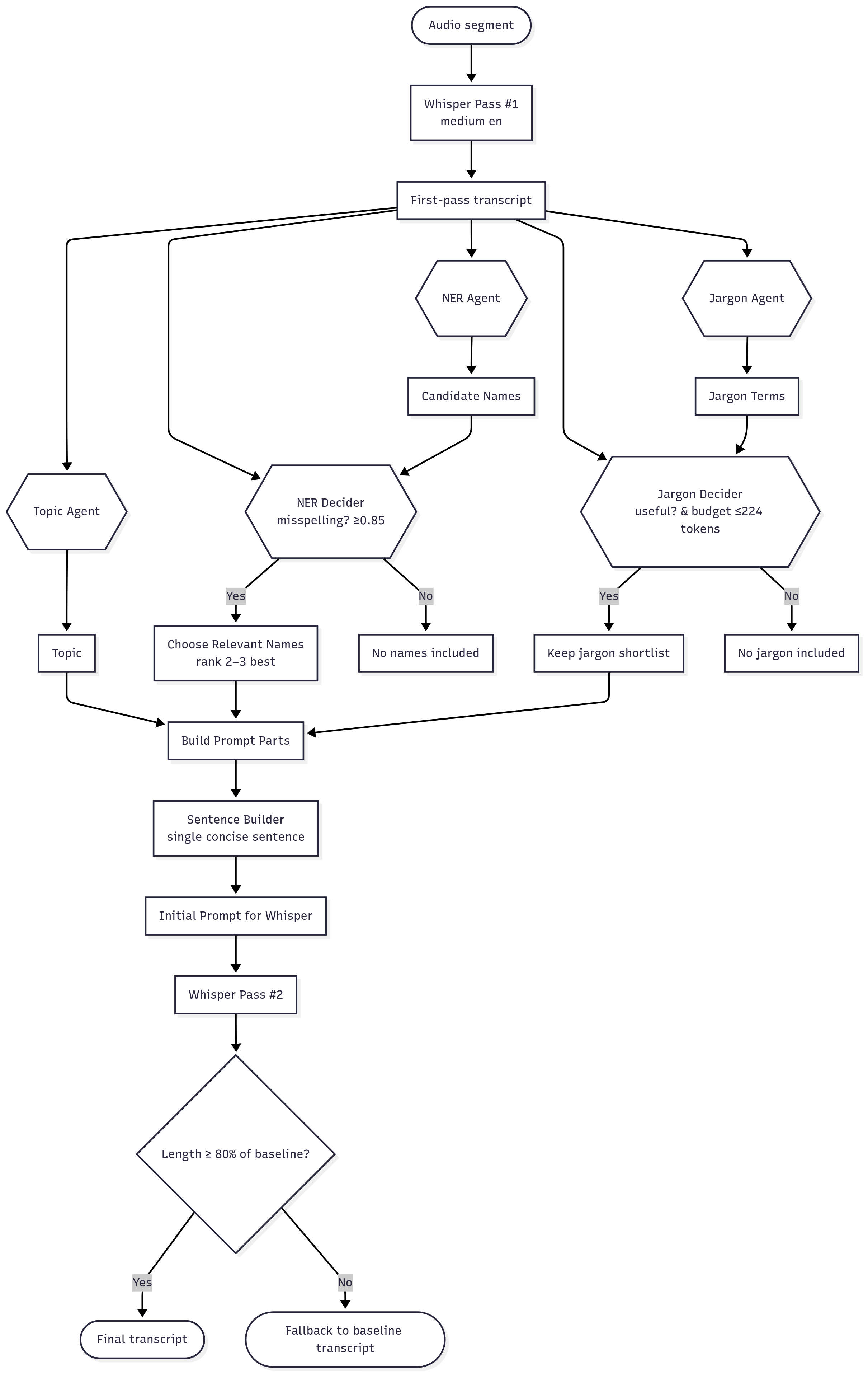}
    \caption{Full multi-agent pipeline flow (P4). Multiple agents extract candidate context from the first-pass transcript. Decider modules filter low-confidence or unnecessary items, and a sentence builder composes a final concise prompt for Whisper’s second pass.}
    \label{fig:myfigure}
\end{figure*}

\textbf{P4 (Full Multi-Agent):} Our most comprehensive pipeline, illustrated in Figure~\ref{fig:myfigure}, incorporates multiple specialized agents (topic, names, jargon) alongside \emph{decider} modules that filter and validate candidate terms. The deciders are crucial: without them, the system often over-corrects by inserting irrelevant names or jargon. By enforcing confidence thresholds (e.g., NER similarity $\geq 0.85$) and prompt-length constraints ($\leq 224$ tokens), the pipeline ensures that only high-value context is included. Finally, a sentence-builder agent assembles the selected topic, names, and jargon into a single concise natural-language sentence, which is injected as the \texttt{initial\_prompt}. This design balances informativeness with compactness, directly exploiting Whisper’s sensitivity to late-position prompt tokens. The result is a robust prompt that maximizes improvements while minimizing degradation, outperforming all other pipelines. \\

\paragraph{Why a natural-sentence prompt?} Instead of a comma-separated list, we render topic, names, and jargon as one concise sentence because: (1) \emph{Fits the model:} Whisper’s decoder continues natural text; coherent syntax stabilizes decoding. (2) \emph{Adds semantics:} Light role/context cues (e.g., "Warriors guard Stephen Curry") reduce homophones and look-alike surname errors. (3) \emph{Uses the window:} Only the last $\le 224$ prompt tokens matter; put rare names/jargon near the end. (4) \emph{Minimizes clutter:} Sentences avoid long lists and punctuation copying; clear boundaries prevent token merges.

\subsection{Implementation Details}

All agents utilize OpenAI's GPT-4o API with carefully engineered prompts for task-specific performance. Whisper (medium.en) provides consistent baseline transcription. Domain knowledge includes comprehensive NBA player rosters (updated through 2024--25 season) and basketball terminology glossaries compiled from official sources. Fuzzy matching employs combined edit distance and Jaro–Winkler similarity with empirically optimized thresholds. Conservative prompt length limits ($\le$ 20 tokens typical) prevent potential degradation from over-long contexts. Confidence-based fallbacks ensure out-of-domain content defaults to baseline transcription.

\section{Experimental Evaluation}

\subsection{Evaluation Framework}

\textbf{Metrics:} Word Error Rate (WER) serves as our primary metric:
\begin{equation}
\text{WER} = \frac{S + D + I}{N}
\end{equation}
where \textit{S}, \textit{D}, \textit{I} represent substitutions, deletions, insertions, and \textit{N} is total word count. We supplement WER with segment-level improvement analysis and statistical significance testing to ensure robust conclusions.

\textbf{Text Normalization:} Standardized preprocessing eliminates formatting artifacts: case normalization, punctuation removal, number standardization, and whitespace regularization. This focuses evaluation on semantic accuracy rather than stylistic differences.

\textbf{Statistical Analysis:} We employ Wilcoxon signed-rank tests for paired comparisons and calculate effect sizes to quantify practical significance beyond statistical significance.

\subsection{Results}

Table 1 presents comprehensive performance statistics across all pipeline variants on our 421-segment evaluation set. \\
\begin{table*}[t]
\centering
\begin{tabular}{@{}lccc@{}}
\toprule
\textbf{Pipeline} & \textbf{Mean WER $\pm$ SD} & \textbf{Improved} & \textbf{Degraded} \\
\midrule
Baseline & 0.217 $\pm$ 0.232 & — & — \\
P1: Topic-Only & 0.238 $\pm$ 0.247 & 88 (20.9\%) & 109 (25.9\%) \\
P2: LLM Post-Fix & 0.217 $\pm$ 0.231 & 81 (19.2\%) & 82 (19.5\%) \\
P3: Names-Enhanced & 0.210 $\pm$ 0.250 & 155 (36.8\%) & 87 (20.7\%) \\
\textbf{P4: Full Multi-Agent} & \textbf{0.180 $\pm$ 0.217} & \textbf{169 (40.1\%)} & \textbf{30 (7.1\%)} \\
\bottomrule
\end{tabular}
\caption{Comprehensive performance comparison across pipeline variants. Significance levels: *\textit{p} < 0.05, ***\textit{p} < 0.001 (Wilcoxon signed-rank test). The full multi-agent pipeline achieves substantial, statistically significant improvements with minimal degradation risk.}
\label{tab:results}
\end{table*}

\textbf{Key Findings:} The full multi-agent pipeline (P4) demonstrates superior performance across all metrics: \\
(i) 17.0\% relative WER reduction (statistically significant) \\
(ii) 40.1\% of segments show improvement vs. only 7.1\% degradation \\
(iii) large effect size indicating practical significance. Notably, the topic-only approach (P1) actually increases WER, demonstrating that broad context without specific corrections can mislead the ASR system. The LLM post-correction approach (P2) shows minimal impact, confirming that prompting-based enhancement significantly outperforms post-processing corrections. The names-enhanced pipeline (P3) achieves moderate improvements but with higher degradation rates, highlighting the importance of decision filtering mechanisms present in P4. \\

\textbf{Degradation:} In some cases, we observed that the initial prompt produced severely degraded results, with sentences truncated or shortened—likely due to suboptimal tokenization. To mitigate this, we implemented a safeguard requiring the output transcript to be at least 80\% the length of the original Whisper transcript before acceptance. This safeguard was triggered in just under 10\% of the evaluated segments, preventing misleadingly shortened outputs from being introduced into the pipeline. While such measures remain necessary given the limitations of current models, we expect that future-generation ASR--LLM systems (e.g., successors such as GPT-5 and beyond) may inherently handle tokenization more robustly, reducing or even eliminating the need for this constraint.

\subsection{Error Analysis and Success Cases}

\textbf{Successful Corrections:} The pipeline excels at:\\
\textit{Proper Noun Recovery:} "anteto kumbo" → "Antetokounmpo", "por zingas" → "Porzingis"\\
\textit{Technical Term Restoration:} "fast brake" → "fast break", "picker roll" → "pick and roll"\\
\textit{Context-Driven Disambiguation:} "the pass to 80" → "the pass to A.D." (Anthony Davis nickname resolution)

\textbf{Failure Modes:} Remaining errors typically involve:\\
\textit{Over-correction (3.2\% of cases):} False positive name detection leading to incorrect insertions.\\
\textit{Out-of-scope Terminology (1.8\% of cases):} Rare nicknames or phrases not in domain glossaries.

\section{Discussion}

\subsection{Practical Applications}

Our approach enables several real-world applications:  \\
(i) \textbf{Live Sports Captioning:} Integration into broadcast pipelines with acceptable latency overhead for enhanced accuracy. \\
(ii) \textbf{Domain-Specific Transcription Services:} Professional services can offer "domain boost" options detecting context and applying relevant terminology. \\
(iii) \textbf{Specialized Voice Assistants:} Medical, legal, or technical environments can route recognition through domain-aware LLM preprocessing. The method's model-agnostic nature avoids vendor lock-in and expensive retraining cycles. \\

\subsection{Computational Overhead Analysis}

Pipeline processing adds 3.3 seconds median latency per segment (acceptable for many applications), with GPT-4o API costs of approximately \$0.008 per segment. For deployment scenarios requiring real-time processing, agent parallelization and local LLM alternatives could reduce latency and costs.

\subsection{Limitations}

Several constraints merit consideration: (i) \textbf{Domain Knowledge Maintenance:} Rosters and terminology evolve, requiring periodic updates. (ii) \textbf{Generalization Uncertainty:} While our architecture is domain-agnostic, validation across medical, legal, or technical domains remains future work. (iii) \textbf{API Dependencies:} Commercial LLM reliance introduces cost and availability considerations, though local alternatives are emerging.

\section{Future Work}

\textbf{Cross-Domain Validation:} Immediate priorities include evaluation on medical transcription, legal depositions, and academic lectures to demonstrate generalizability.
\textbf{Real-Time Optimization:} Investigating agent parallelization, local LLM deployment, and streaming ASR integration for latency-critical applications.
\textbf{Multimodal Enhancement:} Sports commentary could benefit from video context (player recognition, game state) to further reduce ambiguity.
\textbf{Adaptive Learning:} Incorporating user feedback loops to dynamically update domain knowledge and agent behaviors.

\section{Conclusion}

We present a novel multi-agent LLM pipeline for domain-specific ASR enhancement that achieves substantial, statistically significant improvements over baseline Whisper performance. Our approach demonstrates that intelligent prompting can effectively guide ASR systems using external knowledge without expensive model retraining. The 17.0\% relative WER reduction with minimal degradation risk establishes prompt-based enhancement as a viable alternative to traditional domain adaptation approaches. The multi-agent architecture's modularity enables targeted error correction while maintaining system reliability through decision filtering mechanisms. This work opens promising directions for ASR–LLM collaboration, suggesting that hybrid systems leveraging specialized model strengths may become standard for complex AI applications requiring both acoustic modeling and contextual reasoning capabilities.

\clearpage

\onecolumn
\bibliographystyle{plainnat}
\bibliography{references}

\clearpage
\appendix
\section{Prompt Templates and Pipeline Diagram}
\label{appendix:prompts}

This appendix documents the core prompt templates used in our multi-agent ASR enhancement pipeline, and provides a schematic diagram of the pipeline for reproducibility and clarity.

% ----------------------------
% Listings setup
% ----------------------------
\lstset{
    basicstyle=\ttfamily\footnotesize,
    breaklines=true,
    breakatwhitespace=false,
    columns=fullflexible,
    frame=single
}

\subsection{Prompt Templates}

\subsubsection*{Topic Extraction Agent}
\begin{lstlisting}
TOPIC_INSTRUCTIONS = """
Extract the main topic/domain from the transcript.
The output must be between 2--5 words exactly.
"""
\end{lstlisting}

\subsubsection*{Named Entity Recognition (NER) Agent}
\begin{lstlisting}
NER_AGENT_INSTRUCTIONS = """
System:
You receive two inputs--
transcript: full dialogue text.
topic: domain context (e.g., "NBA game").

Task:
1. Extract PERSON entities from the transcript.
2. Load the topic-specific reference list (e.g., basketball roster) and normalize names.
3. For each extracted name:
   a. Compute fuzzy similarity (Damerau--Levenshtein or Levenshtein) against all reference names.
   b. If max similarity >= 0.85, replace with reference name; else, keep the original.
4. De-duplicate, apply proper capitalization/hyphenation, and output names in first-appearance order as a comma-separated string.

Output only the comma-separated list.
"""
\end{lstlisting}

\subsubsection*{Jargon Extraction Agent}
\begin{lstlisting}
JARGON_AGENT_INSTRUCTIONS = """
System:
You receive two inputs--
topic/domain: the subject area for glossary selection.
transcript: the raw commentary text.

Task:
1. Load the domain--specific glossary or ontology.
2. Clean the transcript (remove timestamps/noise, tokenize).
3. Extract candidate jargon via TF--IDF, RAKE, and YAKE.
4. For each candidate, apply Damerau--Levenshtein and SymSpell; if similarity >= 0.90, correct spelling.
5. Include only terms according to the transcript and domain, excluding person names.
6. Deduplicate, filter invalid tokens, apply proper casing.

Output only the comma--separated list.
"""
\end{lstlisting}

\subsubsection*{NER Decision Agent}
\begin{lstlisting}
NER_DECIDER_AGENT_INSTRUCTIONS = """
System:
You are NameMisspellingDetector, a decision-only agent that checks for misspelled person names in a transcript using a provided reference list.

Inputs:
- transcript: full transcript text.
- domain_lexicon: comma-separated list of correct domain names.

Task:
1. Extract PERSON entities via NER.
2. Normalize extracted names and lexicon entries (lowercase, strip punctuation).
3. For each extracted name:
   a. Compute similarity against lexicon entries.
   b. Mark as misspelled if similarity >= 0.85.
4. Answer "YES" if any name is misspelled, else "NO".

Output:
{
  "Answer": "YES" | "NO",
  "Reason": "<brief explanation>"
}
"""
\end{lstlisting}

\subsubsection*{Jargon Decision Agent}
\begin{lstlisting}
JARGON_DECIDER_AGENT_INSTRUCTIONS = """
System:
You are JargonPromptDecider, a decision--only agent that decides whether adding jargon terms to Whisper initial_prompt will improve accuracy.

Inputs:
- transcript: first-pass transcription.
- topic: domain description.
- jargon_list: comma-separated list of jargon terms.

Logic:
1. Normalize text and jargon terms.
2. Detect misspellings using fuzzy matching (>= 0.85).
3. Check prompt budget >= 224 tokens.
4. Answer "YES" if any term is misspelled and budget is within limit.

Output:
{
  "Answer": "YES" | "NO",
  "Reason": "<brief explanation>"
}
"""
\end{lstlisting}

\subsubsection*{Best Candidate Names Agent}
\begin{lstlisting}
BEST_CANDIDATES_AGENT_INSTRUCTIONS = """
System:
You are NameMisspellingRankerAgent, ranking person--name entities by likelihood of being misspelled.

Inputs:
- transcript: first-pass transcription.
- names_list: correct domain names.

Task:
1. Extract PERSON entities via NER.
2. Normalize and compute similarity scores.
3. Rank candidates by misspelling risk.
4. Output top 2--3 corrected names from names list.

Output:
{
  "names": ["<name1>", "<name2>", "<name3>"]
}
"""
\end{lstlisting}

\subsubsection*{Sentence Builder Agent}
\begin{lstlisting}
BUILD_SENTENCE_FROM_PARTS = """
System:
You are JargonNameCombinerAgent, generating a concise sentence with:
- topic
- selected names
- selected jargon

Requirements:
Must include all names from names_list.
Output exactly one fluent sentence.
"""
\end{lstlisting}

\subsubsection*{Transcript Fix Agent}
\begin{lstlisting}
FIX_STT_OUTPUT_AGENT_INSTRUCTIONS = """
You are an expert post-ASR copy-editor.

Task:
Fix obvious ASR errors (spelling, casing, punctuation) without changing meaning or order of lines.

Rules:
Do not merge, split, reorder, or paraphrase lines.
Preserve speaker labels/timestamps.
If unsure, leave token unchanged.

Output corrected transcript only.
"""
\end{lstlisting}
\clearpage
\appendix
\section*{Appendix B: Example Transcriptions}

We present four representative NBA commentary segments where the full multi-agent pipeline (P4) corrects
critical Whisper errors. Each example shows the ground truth, baseline Whisper output,
and the improved P4 transcript with corresponding WER. Corrections of names or jargon are emphasized.

\newenvironment{exampleblock}[1]{%
  \vspace{1em}\noindent\textbf{#1}\par
  \vspace{0.5em}
}{\vspace{1em}}

% ---------------- Example 1 ----------------
\begin{exampleblock}{Example 1 — Name correction: Mispronounced player (Tt8Cepx7HqY\_seg011.wav)}
\begin{tabular}{@{}p{2.8cm}p{11cm}p{1.5cm}@{}}
\toprule
System & Transcript & WER \\
\midrule
Ground Truth & Jackson not a smart pass that time. \textbf{Booker for three. It’s good! Devin Booker from downtown.} It’s a one-point game. & --- \\
Baseline & jackson not a smart pass that time booker for three \emph{devon bucker} from downtown its a onepoint game & 0.20 \\
P4 & jackson not a smart pass that time booker for three \emph{its good devin booker} from downtown its a one point game & 0.10 \\
\bottomrule
\end{tabular}
\end{exampleblock}

% ---------------- Example 2 ----------------
\begin{exampleblock}{Example 2 — Jargon correction: Foul-line jumper (Tt8Cepx7HqY\_seg013.wav)}
\begin{tabular}{@{}p{2.8cm}p{11cm}p{1.5cm}@{}}
\toprule
System & Transcript & WER \\
\midrule
Ground Truth & \textbf{Foul line jumper, Devin Booker, and the foul.} Right now Booker is unguardable. & --- \\
Baseline & \emph{foul on jumper} right now booker is \emph{on guard} & 0.62 \\
P4 & \emph{foul line jumper devin booker and the foul} right now booker is \emph{unguarded} & 0.15 \\
\bottomrule
\end{tabular}
\end{exampleblock}

% ---------------- Example 3 ----------------
\begin{exampleblock}{Example 3 — Full play recovery: Three-pointer sequence (Tt8Cepx7HqY\_seg017.wav)}
\begin{tabular}{@{}p{2.8cm}p{11cm}p{1.5cm}@{}}
\toprule
System & Transcript & WER \\
\midrule
Ground Truth & Rebound. Cousins couldn’t hold on, Booker off-balance gets it. \textbf{A Cameron Payne three-pointer puts it in. Cameron Payne from downtown.} & --- \\
Baseline & rebound cousins couldnt hold on booker off balance gets it \emph{a cameron payne threepointer} & 0.45 \\
P4 & rebound cousins couldnt hold on booker off balance gets it \emph{to cameron payne threepointer puts it in cameron payne from downtown} & 0.15 \\
\bottomrule
\end{tabular}
\end{exampleblock}

% ---------------- Example 4 ----------------
\begin{exampleblock}{Example 4 — Named entities: Tip-off with player names (-71SWHoWQJI\_seg002.wav)}
\begin{tabular}{@{}p{2.8cm}p{11cm}p{1.5cm}@{}}
\toprule
System & Transcript & WER \\
\midrule
Ground Truth & to play game four of round two of the NBA playoffs against the top-seeded Thunder. \textbf{The Holmgren against Jokic at the mid-court circle. Tap is controlled by Jokic.} & --- \\
Baseline & to play game four of round two of the nba playoffs against the top seeded thunder & 0.54 \\
P4 & to play game four of round two of the nba playoffs against the top seeded thunder \emph{holmgren against jokic at the midcourt circle tap is controlled by jokic} & 0.11 \\
\bottomrule
\end{tabular}
\end{exampleblock}

\end{document}